\documentclass{article} %
\usepackage{iclr2023_conference,times}

\iclrfinalcopy

\usepackage[utf8]{inputenc} %
\usepackage[T1]{fontenc}    %
\usepackage[hidelinks]{hyperref}
\hypersetup{
    colorlinks=true,
    linkcolor=blue,
    urlcolor=blue,
    citecolor=blue,
    anchorcolor=blue}
\usepackage{url}            %
\usepackage{booktabs}       %
\usepackage{amsfonts}       %
\usepackage{nicefrac}       %
\usepackage{microtype}      %
\usepackage{xcolor}         %
\usepackage{wrapfig}

\usepackage{graphicx}
\usepackage{subfigure}
\usepackage{booktabs} %

\usepackage{hyperref}
\usepackage{enumitem}

\usepackage[linesnumbered, ruled,vlined]{algorithm2e}
\usepackage{amsthm}
\usepackage{amsmath}
\usepackage{amssymb}
\usepackage{mathtools}
\usepackage{comment}
\usepackage{thmtools,thm-restate}

\definecolor{blue}{HTML}{2B50AA}

\usepackage[capitalize,noabbrev]{cleveref}

\theoremstyle{plain}
\newtheorem{theorem}{Theorem}[section]

\theoremstyle{definition}
\newtheorem{definition}[theorem]{Definition}

\theoremstyle{remark}

\DeclareMathOperator*{\argmax}{arg\,max}

\usepackage[textsize=tiny]{todonotes}

\newcommand{\ps}{\rho} %
\newcommand{\pt}{\nu} %
\newcommand{\Ms}{\mathcal{M}_S}
\newcommand{\Mt}{\mathcal{M}_T}
\newcommand{\Wass}{\mathcal{W}}

\title{Transfer RL via the Undo Maps Formalism}

\author{Abhi Gupta  \\
Massachusetts Institute of Technology\\
\texttt{abhig@mit.edu} \\
\And
Ted Moskovitz \\
University College London\\
\texttt{theodore.moskovitz.19@ucl.ac.uk} \\
\And
David Alvarez-Melis \\
Microsoft Research  \\
\texttt{Alvarez.Melis@microsoft.com}\\
\And
Aldo Pacchiano\\
Microsoft Research, NYC \\
\texttt{apacchiano@microsoft.com}
}

\begin{document}

\maketitle

\begin{abstract}
Transferring knowledge across domains is one of the most fundamental problems in machine learning, but doing so effectively in the context of reinforcement learning remains largely an open problem. Current methods make strong assumptions on the specifics of the task, often lack principled objectives, and---crucially---modify individual policies, which might be sub-optimal when the domains differ due to a drift in the state space, i.e., it is intrinsic to the environment and therefore affects \textit{every} agent interacting with it. To address these drawbacks, we propose \textsf{TvD}: transfer via distribution matching, a framework to transfer knowledge across interactive domains. We approach the problem from a data-centric perspective, characterizing the discrepancy in environments by means of (potentially complex) transformation between their state spaces, and thus posing the problem of transfer as learning to \textit{undo} this transformation. To accomplish this, we introduce a novel optimization objective based on an optimal transport distance between two distributions over trajectories -- those generated by an already-learned policy in the source domain and a learnable pushforward policy in the target domain. We show this objective leads to a policy update scheme reminiscent of imitation learning, and derive an efficient algorithm to implement it. Our experiments in simple gridworlds show that this method yields successful transfer learning across a wide range of environment transformations.

\end{abstract}

\section{Introduction}

Transferring knowledge acquired in one setting to a new one is one of the hallmarks of intelligence and consequently a highly sought-after goal of machine learning methods. Such adaption is crucial both when the task remains the same but the underlying distribution drifts, or when adapting a trained method to a new task. For example, in the context of reinforcement learning (RL; \citep{Sutton:1998}), one might hope that transferring experience acquired from one environment (the `source') to another (the `target') in which certain properties of the world have changed (e.g., dynamics, appearance, etc) would facilitate faster learning in the target than would otherwise be possible. There is ample evidence that human cognition is highly adaptive in this sense. Classic experiments on visuomotor flexibility have demonstrated that subjects made to wear glasses equipped with prisms which displace or even invert the visual field can adapt and even master complicated motor tasks such as riding a bicycle faster than naive acquisition of such skills \citep{harris1965inversion,fernandoruiz2006displacement}. That is, in such experiments humans and animals are able to identify the nature of the transformation between domains and confine the resulting transfer task to one of learning this transformation while preserving the original optimal policy. This effectively reduces the size of the search space of possible solutions, facilitating rapid adaptation.

Inspired by these capabilities, we propose a new framework for transfer reinforcement learning that addresses the challenges described above: Transfer via Distribution Matching (\textsf{TvD}). At its core is a formulation that seeks a mapping (the `undo map') that aligns trajectories of experience across the source and target domains, thus \textit{undoing} the change in the environment responsible for the discrepancy between environments. We consider a distributional alignment, formalized by means of a novel notion of distance between distributions of trajectories which generalizes the standard Wasserstein distance \citep{kantorovich1960wd}. We show that minimizing the dual formulation of this distance yields an update rule reminiscent of policy gradients \citep{williams1992reinforce} with a particular pseudo-reward objective. Compared to other Wasserstein-based imitation learning approaches, our formulation explicitly parameterizes the transformation between domains, which allows us to map additional trajectories at no additional cost, and can be used to adapt multiple policies (e.g., in a multi-agent setting) by solving a single optimization problem. 
\begin{figure*}%
    \centering
    \includegraphics[width=.5\linewidth]{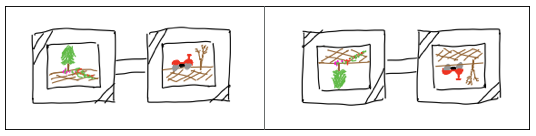}
    \vspace{-8pt}
    \caption{\small \textbf{Motivating example.} The experiment's subjects quickly adapt to the inverted vision field (right).  
    }
    \label{fig:example_first}
    \vspace{-15pt}
\end{figure*}

\section{Background}

\subsection{Reinforcement Learning and Transfer}
In RL, a task is typically modeled as a \textit{Markov decision process} (MDP; \citep{Puterman:2010}) $\mathcal M \triangleq (\mathcal S, \mathcal A, \mathbb P, r, \gamma, d_0)$, where $\mathcal S, \mathcal A$ are (possibly infinite) state and action spaces, $\mathbb P: \mathcal S \times \mathcal A \to \mathcal P(\mathcal S)$ is a transition distribution, $r: \mathcal S \times \mathcal A \to \mathbb R$ is a reward function, $\gamma\in[0, 1)$ is a discount factor, and $d_0\in\mathcal P(\mathcal S)$ is an initial state distribution. The agent takes actions using a stationary policy $\pi: \mathcal S\to\mathcal P(\mathcal A)$ which induces a distribution $\rho^\pi$ over trajectories $\tau \triangleq (s_0, a_0, s_1, a_1, \dots)$, where $\rho^\pi(\tau) = d_0(s_0)\prod_{h\geq 0} \mathbb P(s_{h+1}|s_h, a_h) \pi(a_h|s_h)$. We use $\Gamma$ to denote the space of trajectories. The policy is also associated with a \textit{value} $V^\pi\triangleq \mathbb E_{\tau \sim \rho^\pi} R(\tau)$, where $R(\tau) \triangleq \sum_{h=0}^\infty \gamma^h r(s_h, a_h)$ is called the \textit{return} of the trajectory. The agent's goal is to learn the policy $\pi^\star$ with optimal value: $V^{\pi^\star} \geq V^\pi$ $\forall \pi$. For brevity, we denote $V^\star\triangleq V^{\pi^\star}$ and $\rho^\star\triangleq \rho^{\pi^\star}$. While we consider infinite horizon MDPs here, we note that in practice trajectories are typically truncated to some fixed horizon $H$.

There are a number of specialized problem settings within RL that rely on different assumptions about available information that the agent can leverage in order to speed up learning. In the imitation learning setting \citep{pomerlau1988bc,raychaudhuri2021cdi,stadie2018imitation,papagiannis2020imitation,liu2020State}, we assume access to trajectories $\{\tau_n\}_{n=1}^N \sim \rho^\star$ or more generally, $\{\tau_n\}_{n=1}^N \sim \rho^B$, where $\pi^B$ is a \textit{behavior} policy that the agent seeks to emulate. The objective in this case is find a new policy $\pi$ that minimizes $D(\rho^{\pi}, \rho^B)$, where $D(\cdot, \cdot)$ is some probability metric. In the transfer learning setting \citep{gupta2017transfer,Dadashi2021-ry,BouAmmar_Eaton_Ruvolo_Taylor_2015,liu2022provably}, we consider a source MDP $\Ms = (\mathcal S^S, \mathcal A^S, \mathbb P^S, r^S, \gamma, d^S_0)$ and target MDP $\Mt = (\mathcal S^T, \mathcal A^T, \mathbb P^T, r^T, \gamma, d^T_0)$ with trajectory spaces $\Gamma^S$ and $\Gamma^T$. The agent first interacts with $\Ms$ for example by learning an optimal policy in this domain or building a good model of the environment. Subsequently, the agent interacts with a target domain $\Mt$, where its learning objectives will be made simpler due to the agent's knowledge of information about $\Ms$. For example, the agent may have learned an optimal policy in $\Ms$ with the goal that this policy, once transferred to $\Mt$, will require relatively little fine-tuning to learn the optimal policy for $\Mt$. This is of practical use in many real-world problems, such as training a robot in simulation before deploying it in the real world. The underlying assumption in this case is that there is common structure shared between $\Ms$ and $\Mt$---otherwise, initially training in $\Ms$ would provide no benefit. From this point forward, we denote any trajectory distribution in a source MDP $\Ms$ by $\ps$ and any trajectory distribution in a target MDP $\Mt$ by $\pt$. Similarly, policies in the source domain are denoted by $\pi$ and policies in the target domain are denoted by $\mu$.

\subsection{Distributional Comparison}

In this work we focus on a specific form of task relatedness modulated by what we define as an \textit{undo map}. Although more general definitions of undo maps are possible, for simplicity of exposition, we focus on source and target MDP pairs $(\Ms, \Mt)$  having the same action set $\mathcal{A}$. Note that our assumption of a shared action space is less restrictive than many other multitask RL approaches, which typically assume shared state and action spaces \citep{teh2017distral,Moskovitz:2021_rpotheory}, or even additionally transition kernels \citep{Barreto:2020_fastgpi,moskovitz2022_fr}.

Our formalism seeks to capture the setting where there is an undo map $u_\star : \mathcal{S}^T \rightarrow \mathcal{S}^S$ with $u_\star \in \mathcal{U}$ (a known class of maps $\mathcal{U}$) that transforms states of $\Mt$ into `semantically-equivalent' states of $\Ms$ so that knowledge of the best action to execute in states of $\Ms$ can be lifted to an optimal policy for $\Mt$. In this setting, knowledge of $u_\star$ and an optimal $\Ms$ policy $\pi^\star_S$ yields an optimal policy for $\Mt$ via the composition $\pi^\star_T = \pi^\star_S \circ u_\star$. In other words, knowledge of $\pi^\star_S$ can be used to search over possible value of $u_\star$ to build $\pi^\star_T$. Thus, the hardness of learning $u_\star$ should scale with the size of $\mathcal{U}$. We formalize these intuitions via the following distributional equivalence assumption between the two MDPs $\Ms$ and $\Mt$. 

\begin{definition}[Distributional Equivalence]\label{definition::distributional_equivalence}
We say two MDPs $\Ms$ and $\Mt$ are \emph{distributionally equivalent} if there exists an undo map $u_* : \mathcal{S}^T \rightarrow \mathcal{S}^S$ and a distributional distance $D(\cdot, \cdot)$ (or divergence) in the source domain such that,
\begin{equation}\label{equation::distributional_equivalence}
D( \rho^{\pi_S} , u_\star(  \nu^{ u_\star \circ \pi_S})) = 0,
\end{equation}
for all source policies $\pi_S$ and where $u(p)$ corresponds to the distribution of $u(x)$ with $x\sim p$. When $p$ is a distribution over trajectories $\tau = (s_0, a_0, s_1, a_1, \cdots)$ we define $u(\tau) = (  u(s_0), a_0, u(s_1), a_1, \cdots )$. For any $u \in \mathcal{U}$, $u \circ \pi_S$ is a policy acting in $\Mt$. 
\end{definition}

When $\Mt$ and $\Ms$ are distributionally equivalent, and we have access to $\pi_S$; a natural algorithmic approach to find an approximate version of $\pi_S \circ u_\star$ is to minimize 
\begin{equation}\label{equation::main_objective}
\min_{u \in \mathcal{U}}\  D( \rho^{\pi_S} , u(  \nu^{ u \circ \pi_S}))
\end{equation}
 This insight is the basis of our approach. When $\pi_S = \pi_S^\star$, this distributional objective can be augmented with the rewards of $\Mt$ (if available). In this case the policy optimization objective over $\Mt$ can be written as $\min_u D(\rho^{\pi_S} , u(  \nu^{ u \circ \pi_S})) - \lambda \mathbb{E}_{\tau \sim \nu }\left[ \sum_{(s,a) \in \tau} r^T(s,a) \right]$ for some parameter $\lambda \geq 0$. In all other cases (when $\pi_S$ is arbitrary) the sole distributional objective solves the problem of cross-domain imitation learning, i.e. finding a policy in $\Mt$ that does acts in the target similarly to how $\pi_S$ acts in the source. %

There are many ways to compute distances between distributions $D(\cdot, \cdot)$. In this work we use distributional distances such as the Wasserstein distance and divergences such as $f-$divergences to define the optimization objectives required to solve different transfer learning problems in RL domains. We start by introducing the necessary background to understand their mathematical properties. 

\subsubsection{Optimal Transport Preliminaries}\label{section::optimal_transport}

The Wasserstein distance is a natural way to compute distances between two distributions $p_1, p_2$ with support in sets $\mathcal{X}_1, \mathcal{X}_2$ when we have access to a cost function $c : \mathcal{X}_1 \times \mathcal{X}_2 \rightarrow \mathbb{R} $. The Wasserstein distance $\Wass(p_1, p_2 )$ between distributions $p_1, p_2$ is defined as the minimal expected cost among all couplings between $p_1$ and $p_2$. For the purposes of this work, however, it will be more useful to use its dual formulation,
\begin{equation}\label{equation::dual_formulation}
\begin{split}
&\Wass_c(p_1, p_2) = \max_{h,g \in \mathcal{F}} \left \{ \mathbb{E}_{x \sim p_1}[ h(x) ] + \mathbb{E}_{x \sim p_2}[g(x)] \right\} \\ &\text{s.t.} \quad h(x) + g(x') \leq c(x,x'),~\forall x,x' \in \mathcal{X}_1 \times \mathcal{X}_2,
\end{split}
\end{equation}
where $\mathcal{F}$ is a function class defined by the geometry $c$ induces on $\mathcal{X}_1 \times \mathcal{X}_2$. We will henceforth refer to functions in $\mathcal{F}$ as `test functions.' As an example, when $\mathcal{X}_1, \mathcal{X}_2 \subseteq \mathbb{R}^d$ and $c(x_1, x_2) = \| x_1 - x_2\|$, the set of functions $\mathcal{F}$ in the variational definition of the Wasserstein distance (Equation~\ref{equation::dual_formulation}) corresponds to the set of $1$-Lipschitz functions. 

To compute the Wasserstein distance between two distributions, it is enough to solve the optimization problem in Equation~\ref{equation::dual_formulation}. Due to the constrained nature of this objective, solving this optimization problem can prove challenging. Instead, we'll consider the following regularized version via a soft constraint on the test functions $h$ and $g$,%
\begin{equation}\label{equation::dual_formulation_regularized}
\widetilde{\Wass}_c(p_1, p_2) = \max_{h,g \in \mathcal{F}} \left \{ \mathbb{E}_{x \sim p_1}[ h(x) ] + \mathbb{E}_{x \sim p_2}[g(x)] - \alpha\mathbb{E}_{x, x' \sim p_1 \bigotimes p_2}\left[ \left( h(x) + g(x')  -c(x,x') \right)_{+} \right] \right\},
\end{equation}
where $\alpha >0$ is a regularization parameter and $p_1 \bigotimes p_2$ is the product distribution of $p_1$ and $p_2$ (i.e. i.i.d. samples from each). When $\mathcal{F}$ is a parametric function class, this formulatiom is especially useful because it allows us to compute stochastic gradients of the regularized version (Equation ~\ref{equation::dual_formulation_regularized}). 

Assume $h$ is parametrized by $\xi_1$ and $g$ is parametrized by $\xi_2$ so that $h(x) = F(x,\xi_1)$ and $g(x) = F(x,\xi_2)$. When $\mathcal{F}$ is a class of neural networks, the evaluation function $F(\cdot, \cdot)$ can be thought of as the feed forward operation of input $x$ through the network architecture with parameters $\xi$. Finding an approximation to $h^*,g^*$ can then be found by stochastic gradient ascent on the parameters $\xi_1$ and $\xi_2$, 
\begin{align}
\xi_i &\leftarrow  \xi_i + \eta \left[ \nabla_{\xi_i} F(x_i, \xi_i) - \alpha \nabla_{\xi_i}\left( F(x_1, \xi_1)  - F(x_2, \xi_2) - c(x_1, x_2)\right)_{+} \right], \label{equation::potentials_stochastic_update}
\end{align}
where $ (x_1, x_2) \sim p_1 \bigotimes p_2$ and $\eta$ is a learning rate parameter. A discussion on how to derive similar results for $f-$divergences using their variational form can be found in Appendix~\ref{section::f_divergences_appendix}.

\section{Our Approach}

We focus on approximate ways of minimizing Equation~\ref{equation::main_objective} when $D(\cdot , \cdot)$ is the Wasserstein distance\footnote{The same derivations can be performed using an $f-$divercence instead using the results of Appendix~\ref{section::f_divergences_appendix}.} induced by a per-source-trajectory cost $c : \Gamma^S \times \Gamma^S \rightarrow \mathbb{R}$. In our experiments we define $c$ as the Dynamic Time Warping distance (\textsf{DTW}). See Appendix~\ref{section::dtw} for more information. Assuming access to a parametric form of the undo map class $\mathcal{U}$ where all $u \in \mathcal{U}$ can be written as $u(\cdot) = \mathrm{U}(\cdot, \omega) $ for a parameter $\omega \in \mathbb{R}^{d_U}$. Here $\mathrm{U}$ corresponds to (for example) a class of neural network architectures where $\omega$ equals the parameter choice. For simplicity we'll use the notation $u_\omega$ to denote the undo map parametrized by $\omega$. Provided access to samples from a behavior source policy $\pi_S$, we wish to find an undo map-source policy pair $(u_{\hat{\omega}}, \pi^{\hat{\theta}}_{S})$ parametrized by $\omega \in \mathbb{R}^{d_U}$ and $\theta \in \mathbb{R}^d$ such that
\begin{equation}\label{equation::wass_objective}
\hat{\omega}, \hat{\theta} = \min_{\omega, \theta} \  \Wass_c(\rho^{\pi_S}, u_\omega\left(  \nu^{u_\omega \circ \pi_S^{\theta}}\right) ) 
\end{equation}
When a parametric form of $\pi_S$ is known, we can restrict ourselves to the simpler objective of learning the undo map only while fixing $\pi_S^{\theta} = \pi_S$ in Equation~\ref{equation::wass_objective}. In our experiments (see Section~\ref{section::experiments}) we explore both of these options and provide the reader with a detailed account of the advantages, disadvantages and limitations of each approach. %

In order to turn Equation~\ref{equation::wass_objective} into a tractable algorithm we make the simplifying assumption that we have access to a parametric form of the dual potentials class for $\Wass_c$ (recall the dual formulation of the Wasserstein distance in Equation~\ref{equation::dual_formulation}) and substitute the unregularized Wasserstein distance with its regularized counterpart (Equation~\ref{equation::dual_formulation_regularized}) .

When the cost function $c$ acts on the space of source trajectories these assumptions naturally lead to the following min-max instantiation of objective~\ref{equation::wass_objective},
\begin{align}
 \min_{\omega, \theta} \max_{\xi_1, \xi_3 } &\left \{ \mathbb{E}_{\tau \sim \rho^{\pi_S}}[ F(\tau, \xi_1) ] + \mathbb{E}_{\tau \sim \nu^{u_\omega \circ \pi_S^\theta} }[F(u_\omega(\tau), \xi_2)] \right.\notag \\
 &\left.- \alpha\mathbb{E}_{\tau, \tau' \sim \rho^{\pi_S} \bigotimes \nu^{u_\omega \circ \pi_S^\theta} }\left[ \left( F(\tau, \xi_1) + F(u_\omega(\tau'), \xi_2)  -c(\tau,u_\omega(\tau')) \right)_{+} \right] \right\}. \label{equation::dual_formulation_regularized_policy}
\end{align}

When the cost function is instead defined only between elements of the state space, the general form of the objective remains similar but instead the $\tau$ symbols are substituted by state variables $s$ and $\rho^{\pi_S},\nu^{u_\omega \circ \pi_S^\theta}$ denote state visitation instead of trajectory distributions.

 Let $L(\omega, \theta,  \xi_1, \xi_2)$ denote the inner objective of Equation~\ref{equation::dual_formulation_regularized_policy}. Our algorithm alternates between optimizing the inner objective of~\ref{equation::dual_formulation_regularized_policy} to produce candidate dual potential parameters $\hat{\xi}_1, \hat{\xi}_2$ and using these potentials to compute stochastic gradient steps on $\theta, \xi$.

 Both $\nabla_{\xi_1, \xi_2} L(\omega, \theta, \xi_1, \xi_2) $ and $\nabla_{\omega, \theta} L(\omega, \theta, \xi_1, \xi_2) $ can be estimated using sample trajectories from the product distribution $\rho^{\pi_S} \bigotimes \nu^{u_\omega \circ \pi_S^{\theta}}$. Their formulas can be derived by treating the dual potentials as 'reward' functions (see equations~\ref{equation::gradient_theta_objective} and~\ref{equation::gradient_omega_objective} in Appendix~\ref{section::gradient_computation}). When the cost function is defined between source MDP states, the formulas in~\ref{equation::gradient_theta_objective} and~\ref{equation::gradient_omega_objective} attain the same form with trajectory samples substituted by state samples from the state visitation distribution of policies $\pi_S$ and $u_\omega \circ \pi_S^\theta$. The inner optimization over parameters $\xi_1, \xi_2$ can be done by taking stochastic gradient steps. The resulting gradient steps in the \textsf{TvD} algorithm are intimately related to the Behavior Guided policy optimization paradigm of~\citep{pacchiano2020learning,moskovitz2020efficient}.

\begin{algorithm}[h]
\textbf{Input: }  Source policy $\pi_{S}$ (or batch of source domain $\pi_S$ trajectories). \\
\textbf{Initialize: } Initialize policy, undo and potential parameters $\theta^{(0)}, \omega^{(0)}, \xi^{(0)}_1, \xi_2^{(0)}$.\\
\For{episode $t=1, \cdots, N$}{
    Use stochastic gradients (Eq~\ref{equation::potentials_stochastic_update}) to compute approximately optimal potential parameters $$\xi_1^{(t)}, \xi_2^{(t)} = \argmax_{\xi_1, \xi_2} L(\omega^{(t-1)}, \theta^{(t-1)}, \xi_1, \xi_2).$$ \\
    SGD $L$ using approximate versions of $\nabla_{\omega, \theta} L(\omega, \theta, \xi_1^{(t)}, \xi_2^{(2})$ (see equations~\ref{equation::gradient_theta_objective} and~\ref{equation::gradient_omega_objective} ). 
}
\caption{Cross Domain \textbf{T}ransfer \textbf{v}ia \textbf{D}istribution Matching (\textsf{TvD}).}
\label{alg:wassDomainAdaptation}
\end{algorithm}

\section{Experiments}\label{section::experiments}

\subsection{Reasoning in Grid World}
In order to understand the properties of Algorithm \ref{alg:wassDomainAdaptation}, we first consider the navigation task of reaching the bottom-right corner of an 8-by-8 grid. In this grid world, we represent the state as the $(x,y)$ position of the agent where $(0,0)$ denotes its initial location. At any position in the grid, there are four available actions: move left, move right, move up, and move down. The agent always moves to the immediate cell in the direction of the chosen action unless it runs into a wall, in which case the agent remains at the same place. An episode terminates either when the agent reaches the goal $(7,7)$ or $H=50$ timesteps have passed. Finally, we provide a per-step reward of $-1$ in order to encourage the agent to reach the destination as quickly as possible. 

From this source MDP $\Ms$, we define a related but different target MDP $\Mt$ by considering transformations to the state space of $\Ms$. In particular, we rotate the grid world of the original domain so in $\Mt$ every position  $\binom{x}{y}$ appears to be $T_\theta \binom{x}{y}$ instead, where $T_\theta \in \mathbb{R}^{2\times2}$ is a rotation matrix parameterized by a single angle $\theta$. Although we consider a variety of angles, we only visualize experiments for $\theta=\frac{\pi}{2}$ because they are much easier to interpret. In this setting, the optimal undo map $u_\star$ rotates the grid world defined in $\Mt$ by $-\theta$ in order to undo the transformation between the two domains, therefore $u_\star\binom{x}{y} = T_\theta^{-1}\binom{x}{y}$. Since $T_\theta$ is an orthonormal matrix, the inverse $T_\theta^{-1}$ is equivalent to $T_\theta^\top$. We can easily observe that for any policy $\pi$ acting in the source domain, rolling out the the policy $T_\theta^{-1} \circ \pi$ in the target domain produces a trajectory distribution which transformed with $T_\theta^{-1}$ matches the trajectory distribution of $\pi$ in the source. 

\begin{figure*}[t]
    \centering
    \includegraphics[width=.8\linewidth]{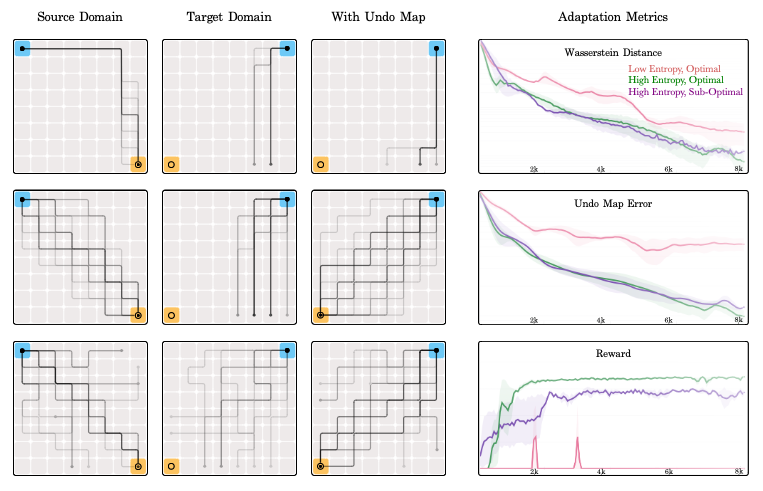}
    \vspace{-8pt}
    \caption{\small \textbf{Policy Behavior with and without Undo Maps} We visualize the trajectory distribution of three different policies when deployed in the source domain where they were trained, when re-used in a new target domain whose state space has drifted from the source, and when adapted in this target domain through the means of a parametric, undo map. Each row corresponds to one of the three policies whose behavior is characterized by whether it reaches the goal in the grid world (i.e optimal or sub-optimal) and the diversity of its trajectories (i.e. low-entropy or high-entropy). }
    \label{fig:example}
    \vspace{-15pt}
\end{figure*}

\subsection{Interacting with the Source and Target Domains}
We assume knowledge of solving the task in $\Ms$ in the form of a parametric policy $\pi_{\theta}$. Although there are many policies that solve the task of reaching the destination in the grid, each one may take a different path. We specifically consider three kinds of interaction with the source domain:

\begin{enumerate}
  \item Low-Entropy, Optimal Behavior: The agent tends to take the same route to the destination, reaching there as quickly as possible but only ever visiting a few of the cells in the grid. We train a low-entropy, optimal policy in the source domain with PPO \citep{Schulman:2017}. 
  \item High-Entropy, Optimal Behavior: The agent takes a variety of paths to reach the destination, visiting a large fraction of the cells in the grid. Although each route is different, the agent still reaches the destination in as few steps as possible. We train a high-entropy, optimal policy in the source domain with a discrete formulation of SAC \citep{haarnoja_2018sac}.
  \item High-Entropy, Sub-optimal Behavior: The agent again visits many cells in the grid, though this time it does not necessarily reach the destination as quickly as possible. At times, the agent even fails to reach the destination and instead runs into a wall. In order to recover a high-entropy, sub-optimal policy, we train a policy in the source domain with a discrete formulation of SAC for only a few epochs. 
\end{enumerate}

We visualize each of the behaviors described above by considering our grid world setting. Figure \ref{fig:example} depicts three columns and three rows, where the first column shows the trajectory distribution of a policy $\pi_{\theta}$ trained in the source domain, the second column shows the trajectory distribution of the same policy now acting in the target domain, and the last column shows the trajectory distribution of the policy $u_{\omega}\circ \pi_{\theta}$. Each row corresponds to a different policy $\pi_{\theta}$ in the source domain trained to demonstrate either low-entropy and optimal behavior, high-entropy and optimal behavior, or high-entropy and sub-optimal behavior. A path or trajectory in any of the plots in Figure \ref{fig:example} is represented as a line starting at the initial state, highlighted in blue, and ending possibly at the destination, highlighted in yellow, or another cell. Frequently occurring trajectories appear in a darker shade than others. 

In the source domain, trajectories from a low-entropy, optimal policy tend to be the same, always moving along the edges of the grid, while trajectories from a high-entropy, optimal policy all end at the bottom-right corner of the grid in many different ways. In contrast, trajectories from a high-entropy, sub-optimal policy look chaotic, often times ending at a cell other than the destination. In the target domain, we observe how a policy trained originally in the source domain behaves when placed in the target domain. Note that the grid in the target domain is rotated by $90^{o}$ from the grid in the source domain. Regardless of the behavior in the source domain, we see that the same policy never reaches the destination. In other words, every policy in the source domain fails to solve the task even once in the target domain. 

\subsection{Adapting with the Undo Map}
We evaluate how well each of the policies $\pi_\theta$ trained in the source domain $\Ms$ can be adapted to the target domain $\Mt$ with the undo map. Following the procedure in Algorithm \ref{alg:wassDomainAdaptation}, our agent acts in the rotated grid world according to the policy $a \sim \pi_{\theta}(\cdot|u_{\omega}(s))$. As before, we visualize the trajectory distribution of $u_{\omega}\circ \pi_{\theta}$ in the last column of Figure \ref{fig:example}. We also track three important metrics over the course of adaptation: the Wasserstein distance $\Wass_c(\rho^{\pi_\theta}, u_\omega\left( \nu^{u_\omega \circ \pi_S^{\theta}}\right))$ estimated from sample trajectories in the source and target domains, the return achieved in the target domain $\mathbb{E}_{\tau \sim \nu }\left[ R(\tau)\right]$ when acting with the policy $u_{\omega}\circ \pi_{\theta}$, and finally the undo map error defined below:

\begin{align}
\mathbb{E}_{(x,y) \sim \rho^{\pi_{\theta}}}{\left[\Big|\Big|\binom{x}{y}-u_{\omega}\circ T_{\theta}\binom{x}{y}\Big|\Big|^{2}\right]}
\end{align}

Here, the distribution $\rho^{\pi_\theta}$ refers to the state visitation frequency or occupancy measure induced in the source domain when acting with $\pi_\theta$. Intuitively, the undo map error measures how close the composition $u_{\omega}\circ T_{\theta}$ is to the identity function provided that we have knowledge of $T_{\theta}$, the transformation applied to the state space of the source domain in order to construct the target domain. Note that we do not use $T_{\theta}$ during learning but rather only as a performance metric when possible. 

In our experiments, we find that it is quite difficult to recover the optimal undo map from a low-entropy, optimal policy in the source domain (plots shown in pink). Although the Wasserstein distance indeed decreases during adaptation, the undo map error plateaus and the agent never learns to reach the destination. This is shown in the last grid of the first row of Figure \ref{fig:example} where trajectories in the target domain after using the undo map always run into a wall. Surprisingly, we find that training the undo map proves challenging only when the source policy has narrow state coverage. For instance, we are able to learn the undo map very well in the case where the source policy has high entropy, whether the behavior is optimal (plots shown in green) or sub-optimal (plots shown in purple). This reiterates the idea that the undo map depends on the source and target domains rather than an already trained, optimal policy in the original environment. A policy with enough state coverage in the source domain can be sufficient to learn the undo map even if the policy is not optimal. After learning the undo map $u_{\omega}$ in this way, we can later compose it with an optimal policy $\pi_{\phi}$ in the source domain to construct an optimal policy  $u_{\omega}\circ \pi_{\phi}$ in the target domain. 

\subsection{Learning from Demonstrations}
We have so far considered transfer settings where knowledge about the source domain is available as a parametric policy $\pi_{\theta}$. In the case that only expert demonstrations $\{\tau_{i}\}_{i=0}^{N}$ from the source domain are available, we can again follow the procedure in Algorithm $1$ with the exception that we must now learn the source policy $\pi_{\theta}$ in addition to the undo map $u_{\omega}$. This works well even when the demonstrations only cover a small portion of the state space. We show this in the same grid world setting by taking demonstrations from a deterministic source policy that moves along the edges of the grid.

When there is no drift in the state spaces of the two domains or the undo map is already known, we can think of this setting as no different than that of imitation learning. The objective is now to simply recover the source policy. Our algorithm in this case reduces exactly to variants of GAIL \citep{ho2016gail}. With these grid world experiments, we come to the conclusion that our algorithm has two principle use cases: solving the same task in a new domain from only demonstrations in the original domain or, more importantly, learning a task-agnostic undo map which allows for reuse but requires high state coverage experts. 

\section{Discussion}

Although the experimental example we have described seemingly requires the assumption that $u_*$ is a map between target and source raw observations, the source domain may be instead an MDP of abstract states, and the undo map a class of mappings between observations of the target and these abstractions. An example of this setting may arise when we train a policy in a source domain where we learn an abstraction (representation) map
When training in the target domain, the mapping between observations and the learned representation from the source may have changed but still remain in a small neighborhood of where the old representation mapping was. The policy to imitate can be thought as a mapping between source domain representations (latent states) and actions. Having access to this, the learner is only required to learn the correct representation map to recover a semantically equivalent behavior policy in the target domain. Once the map from target observations to latent states is known, the optimal policy of the target domain can be read as a composition between this map and the optimal source policy.

It should also be noted that the distributional equivalence condition does not require absolute equality between the `undone distribution' of the target and the behavior policy in the source domain, but only that source and target transformed distributions be close under an appropriate distributional distance. This allows us to avoid assuming the underlying dynamics (for example in abstraction space) of the source and target are exactly the same. By imposing a trajectory based distributional equivalence based on the dynamic time Warping distance, we allow for similarities between target and source domains where the dynamics are different but there is a semantic map between target and source trajectories.

In summary, we introduced a novel optimization objective based on the Wasserstein distance (or an $f-$divergence) between the trajectory distribution induced by an policy in the source domain and that of a a learnable pushforward policy in the target domain. We showed this objective leads to a policy update scheme reminiscent of imitation learning, and derive \textsc{TvD}, an efficient algorithm to implement it. Simple experiments demonstrate that \textsc{TvD} facilitates transfer learning across several environment transformations. In the future, we'd like to scale \textsc{TvD} to more challenging domains and applications.

\bibliography{references}
\bibliographystyle{iclr2023_conference}

\section{Appendix}

\subsubsection{$f-$Divergences}\label{section::f_divergences_appendix}

Given two measures $p_1$ and $p_2$ with support in (shared) $\mathcal{X}$ such that $p_1$ is absolutely continuous in reference to $p_2$ (i.e. $p_1 \ll p_2$), and a convex function $f: (0, \infty) \rightarrow \mathbb{R}$, we define the $f-$divergence between $p_1$ and $p_2$ as
\begin{equation}\label{equation::f_divergence_primal_definition}
D_f( p_1 \parallel p_2) = \mathbb{E}_{x \sim p_2}\left[ f\left( \frac{p_1(x)}{p_2(x)}\right) \right]
\end{equation}
The $f-$divergence $D_f(p_1\parallel p_2)$ can be written in variational form in terms of $f^* : \mathbb{R}\rightarrow \mathbb{R}$, the convex conjugate of $f$ as,
\begin{equation}\label{equation::f_divergence_variational_definition}
D_f( p_1 \parallel p_2) = \sup_{g : \mathcal{X} \rightarrow \mathbb{R}} \mathbb{E}_{x \sim p_1}\left[ g(x)\right] - \mathbb{E}_{x \sim p_2}\left[ f^*(g(x))\right].
\end{equation}
where $f^*(y) = \sup_{z \in (0, \infty)} zy - f(z)$ and $g$ iterates over the set of functions such that both expectations on the right hand side of Equation~\ref{equation::f_divergence_variational_definition} are finite (see~\cite{wu}). Using formula~\ref{equation::f_divergence_variational_definition} we can derive the following variational formulas for the $\chi^2-$divergence, total variation distance (TV) and the Kullback-Leibler (KL) divergence,
\begin{align*}
\chi^2\left( p_1 \parallel p_2 \right) &= \sup_{g : \mathcal{X} \rightarrow \mathbb{R}} \mathbb{E}_{x \sim p_1}\left[ g(x) \right] - \mathbb{E}_{x \sim p_2} \left[ g(x) + \frac{g^2(x)}{4}\right] \\
\mathrm{TV}(  p_1 \parallel  p_2) &= \sup_{g : \mathcal{X} \rightarrow \mathbb{R} \text{ s.t } |g| \leq \frac{1}{2}} \mathbb{E}_{x \sim p_1}\left[ g(x) \right] - \mathbb{E}_{x \sim p_2} \left[ g(x) \right] \\
\mathrm{KL}(p_1 \parallel p_2) &=  \sup_{g : \mathcal{X} \rightarrow \mathbb{R} } \mathbb{E}_{x \sim p_1}\left[ g(x) \right] - \mathbb{E}_{x \sim p_2} \left[ e^{g(x)} \right]
\end{align*}

One limitation of $f$-divergences is the requirement that $p_1 \ll p_2$. Although this is avoidable for the case of the total variation distance, it is required for $\xi^2$ and $\mathrm{KL}$. Just as we did for the Wasserstein gradient objective of Equation~\ref{equation::potentials_stochastic_update} we consider a parametrized form of the objective so that $g(x) = F(x, \xi)$. Finding an approximation to $g^*$ (the maximizer of equation~\ref{equation::f_divergence_variational_definition} when achievable) can be found by stochastic gradient ascent on the parameter $\xi$,
\begin{align}
\xi &\leftarrow  \xi + \eta \left[ \nabla_{\xi} F(x_1, \xi) -  \nabla_{\xi} f^*(F(x_2, \xi))\right], \label{equation::potentials_stochastic_update_f_divergence}
\end{align}
Where $ (x_1, x_2) \sim p_1 \bigotimes p_2$ and $\eta$ is a learning rate parameter. In the case of $\mathrm{TV}$, $\chi^2$ and $\mathrm{KL}$ the stochastic gradients take the form,
\begin{align*}
\chi^2 \rightarrow& \qquad \nabla_{\xi} F(x_1, \xi) -  \nabla_{\xi} f^*(F(x_2, \xi)) = \nabla F(x_1, \xi)  - \left(1+ \frac{F(x_2, \xi)}{2}\right) \nabla F(x_2, \xi)  \\
\mathrm{TV} \rightarrow& \qquad \nabla_{\xi} F(x_1, \xi) -  \nabla_{\xi} f^*(F(x_2, \xi)) =   \nabla F(x_1, \xi)  -  \nabla F(x_2, \xi)\\
\mathrm{KL} \rightarrow& \qquad \nabla_{\xi} F(x_1, \xi) -  \nabla_{\xi} f^*(F(x_2, \xi)) = \nabla F(x_1, \xi) - e^{F(x_2, \xi)} \nabla F(x_2, \xi).
\end{align*}

\section{Gradient Computations}\label{section::gradient_computation}
\subsection{Wasserstein Distance Objective}

 Given a pair $\omega, \theta$ we define $\xi_1(\omega, \theta), \xi_2(\omega, \theta) = \arg\max_{\xi_1, \xi_2} L(\omega, \theta, \xi_1, \xi_2)$. Danskin's theorem implies $\nabla_{\omega, \theta} \max_{\xi_1, \xi_2} L(\omega, \theta, \xi_1, \xi_2) = \nabla_{\omega, \theta} L(\omega, \theta,\xi_1(\omega, \theta), \xi_2(\omega, \theta) )$. Moreover, for any fixed $\xi_1, \xi_2$ the gradient $\nabla_{\omega, \theta} L(\omega, \theta, \xi_1, \xi_2)$ satisfies, 
\begin{align}
\nabla_{\theta} L(\omega, \theta, \xi_1, \xi_2) &=  \mathbb{E}_{\tau \sim \nu^{u_\omega \circ \pi_S^\theta} }\left[ F(u_\omega(\tau), \xi_2)  \nabla_\theta \log(\pi_S^\theta(u_\omega(\tau)))\right]\label{equation::gradient_theta_objective} \\
&\quad - \alpha\mathbb{E}_{\tau, \tau' \sim \rho^{\pi_S} \bigotimes \nu^{u_\omega \circ \pi_S^\theta} }\left[       G(\tau, \tau', \omega, \theta, \xi_1, \xi_2) \nabla_\theta \log(\pi_S^\theta(u_\omega(\tau'))) \right] \notag
\end{align}
Where $\nabla_\theta \log(\pi_S^\theta(u_\omega(\tau))) = \sum_{s_i,a_i \in \tau}\nabla_{\theta} \log\left( \pi_S^\theta( u_\omega(s_i), a_i) \right) $ and $G(\tau, \tau', \omega, \theta, \xi_1, \xi_2) = \left( F(\tau, \xi_1) + F(u_\omega(\tau'), \xi_2)  -c(\tau,u_\omega(\tau')) \right)_{+} $. Similarly,

\begin{align}
\nabla_{\omega} L(\omega, \theta, \xi_1, \xi_2) =  \mathbb{E}_{\tau \sim \nu^{u_\omega \circ \pi_S^\theta} }\left[ F(u_\omega(\tau), \xi_2)  \nabla_\omega \log(\pi_S^\theta(u_\omega(\tau))) + \nabla_\omega F(u_\omega(\tau), \xi_2)\right] \qquad  \label{equation::gradient_omega_objective} \\
- \alpha\mathbb{E}_{\tau, \tau' \sim \rho^{\pi_S} \bigotimes \nu^{u_\omega \circ \pi_S^\theta} }\left[ G(\tau, \tau', \omega, \theta, \xi_1, \xi_2) \nabla_\theta \log(\pi_S^\theta(u_\omega(\tau'))) + \nabla_\theta G(\tau, \tau', \omega, \theta, \xi_1, \xi_2) \right] \notag
\end{align}

Sample versions of these gradient formulas can computed via samples from $\pi_S$ in the source domain and $u_\omega \circ \pi_S$ in the target domain. Algorithm~\ref{alg:wassDomainAdaptation} remains unchanged.

\subsection{$f-$divergence objective}

We can derive a similar expression as Equation~\ref{equation::dual_formulation_regularized_policy} for the setting where we use an $f$-divergence. In this case (using the variational formulation from Equation~\ref{equation::f_divergence_variational_definition}) the min-max objective takes the form,
\begin{align}
 \min_{\omega, \theta} \max_{\xi } &\left \{ \mathbb{E}_{\tau \sim \rho^{\pi_S}}[ F(\tau, \xi) ] - \mathbb{E}_{\tau \sim \nu^{u_\omega \circ \pi_S^\theta} }[f^\star(F(u_\omega(\tau), \xi))] \right\}. \label{equation::dual_formulation_regularized_policy_f_div}
\end{align}

Where we have assumed the function $g( \cdot )$ from Equation~\ref{equation::f_divergence_variational_definition} can be parameterized by parameter $\xi$ as $F(\cdot, \xi)$. If we define as $L(\omega, \theta, \xi)$ as the inner objective of equation~\ref{equation::dual_formulation_regularized_policy_f_div}, the gradients $\nabla_{\omega} L(\omega, \theta, \xi)$ and $\nabla_\theta L(\omega, \theta, \xi)$ satisfy,

\begin{align*}
\nabla_{\theta} L(\omega, \theta, \xi) &=  -\mathbb{E}_{\tau \sim \nu^{u_\omega \circ \pi_S^\theta} }\left[ f^\star(F(u_\omega(\tau), \xi)  )\nabla_\theta \log(\pi_S^\theta(u_\omega(\tau)))\right]\\
\nabla_{\omega} L(\omega, \theta, \xi) &=  -\mathbb{E}_{\tau \sim \nu^{u_\omega \circ \pi_S^\theta} }\left[ f^\star\left(F(u_\omega(\tau), \xi)\right)  \nabla_\omega \log(\pi_S^\theta(u_\omega(\tau))) + \nabla_\omega f^\star(F(u_\omega(\tau), \xi))\right]
\end{align*}
\begin{wrapfigure}{r}{.3\textwidth}

    \centering
    \includegraphics[width=\linewidth]{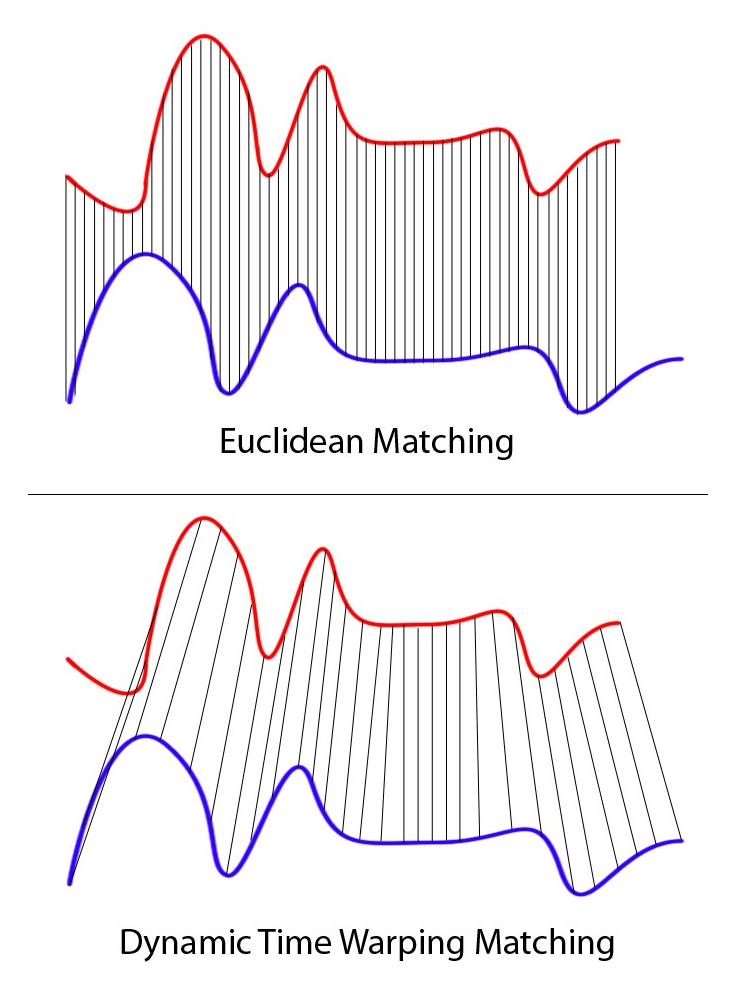}
    \vspace{-8pt}
    \caption{\small \textbf{ Visual representation of the difference between the euclidean distance and the dynamic time warping distance for computing the distance between two time series. (Figure from~\citep{dtwWebsite}).}}
    \label{fig:dtw}
\end{wrapfigure}

Same as in the case of Wasserstein distances, sample versions of these gradient formulas can computed via samples from $\pi_S$ in the source domain and $u_\omega \circ \pi_S$ in the target domain.

\section{Dynamic Time Warping Distance}\label{section::dtw}

We use the Dynamic Time Warping Distance (\textsf{DTW}). This distance between two time series of possibly different sizes is designed to measure their similarity even if they may vary in speed. A detailed description may be found here:~\url{https://en.wikipedia.org/wiki/Dynamic_time_warping}. In our experiments we define the \textsf{DTW} distance between two trajectories with a base per state distance equal to the euclidean $\ell_2$ distance.

\end{document}